\documentclass[sigconf]{acmart}

\usepackage{array}
\usepackage{stfloats}
\usepackage{url}
\usepackage{verbatim}
\usepackage{bbding}
\usepackage{graphicx}
\usepackage{textcomp}
\usepackage{xcolor}
\usepackage{multirow}
\usepackage{booktabs}
\usepackage{makecell}
\usepackage[ruled]{algorithm2e}
\newcommand{\eg}{\textit{e.g.}}
\newcommand{\ie}{\textit{i.e.}}
\newcommand{\our}{GraphControl}
\newcommand{\ours}{GraphControl\space}

\newsavebox\CBox
\def\textBF#1{\sbox\CBox{#1}\resizebox{\wd\CBox}{\ht\CBox}{\textbf{#1}}}
\definecolor{dark2green}{rgb}{0.1, 0.65, 0.3}
\definecolor{dark2orange}{rgb}{0.9, 0.4, 0.}
\definecolor{dark2purple}{rgb}{0.4, 0.4, 0.8}

\DeclareMathOperator*{\argmin}{arg\,min}

\usepackage{dsfont}

\usepackage{hyperref}
\usepackage{tablefootnote}
\usepackage{colortbl}
\definecolor{LightCyan}{rgb}{0.88,1,1}
\definecolor{Gray}{gray}{0.92}

\AtBeginDocument{%
  \providecommand\BibTeX{{%
    \normalfont B\kern-0.5em{\scshape i\kern-0.25em b}\kern-0.8em\TeX}}}

\setcopyright{acmcopyright}
\copyrightyear{2024} \acmYear{2024} \setcopyright{acmlicensed}\acmConference[WWW '24]{Proceedings of the ACM Web Conference 2024}{May 13--17, 2024}{Singapore, Singapore} \acmBooktitle{Proceedings of the ACM Web Conference 2024 (WWW '24), May 13--17, 2024, Singapore, Singapore} 
\acmDOI{10.1145/3589334.3645439} \acmISBN{979-8-4007-0171-9/24/05}

\begin{document}
\title{GraphControl: Adding Conditional Control to Universal Graph Pre-trained Models for Graph Domain Transfer Learning}


\author{Yun Zhu}
\orcid{0000-0002-8950-383X}
\authornote{Equal contributions.}
\affiliation{%
  \institution{Zhejiang University}
  \city{Hangzhou}
  \country{China}
  }
\email{zhuyun_dcd@zju.edu.cn}

\author{Yaoke Wang}
\authornotemark[1]
\affiliation{%
  \institution{Zhejiang University}
  \city{Hangzhou}
  \country{China}
}
\email{wangyaoke@zju.edu.cn}

\author{Haizhou Shi}
\affiliation{%
 \institution{Rutgers University}
 \city{New Brunswick}
 \state{New Jersey}
 \country{US}
 }
 \email{haizhou.shi@rutgers.edu}

\author{Zhenshuo Zhang}
\affiliation{%
  \institution{Zhejiang University}
  \city{Hangzhou}
  \country{China}
  }
\email{zs.zhang@zju.edu.cn}

\author{Dian Jiao}
\affiliation{
  \institution{Zhejiang University}
  \city{Hangzhou}
  \country{China}
  }
\email{jd_dcd@zju.edu.cn}

\author{Siliang Tang}
\authornote{Corresponding author.}
\affiliation{%
  \institution{Zhejiang University}
  \city{Hangzhou}
  \country{China}
  }
\email{siliang@zju.edu.cn}

\renewcommand{\shortauthors}{Yun Zhu, et al.}


\begin{abstract}
Graph self-supervised algorithms have achieved significant success in acquiring generic knowledge from abundant unlabeled graph data. These pre-trained models can be applied to various downstream Web applications, saving training time and improving downstream performance.
However, variations in attribute semantics across graphs pose challenges in transferring pre-trained models to downstream tasks. Concretely speaking, for example, the additional task-specific node information in downstream tasks (specificity) is usually deliberately omitted so that the pre-trained representation (transferability) can be leveraged. The trade-off as such is termed as ``transferability-specificity dilemma'' in this work.
To address this challenge, we introduce an innovative deployment module coined as GraphControl, motivated by ControlNet, to realize better graph domain transfer learning.
Specifically, by leveraging universal structural pre-trained models and GraphControl, we align the input space across various graphs and incorporate unique characteristics of target data as conditional inputs. 
These conditions will be progressively integrated into the model during fine-tuning or prompt tuning through ControlNet, facilitating personalized deployment.
Extensive experiments show that our method significantly enhances the adaptability of pre-trained models on target attributed datasets, achieving 1.4-3x performance gain. Furthermore, it outperforms training-from-scratch methods on target data with a comparable margin and exhibits faster convergence. Our codes are available at: \href{https://github.com/wykk00/GraphControl}{\texttt{https://github.com/wykk00/GraphControl}}
\end{abstract}




\maketitle
\sloppy
\section{Introduction}
Graph-structured data is prevalent in Web applications, including community detection~\cite{community}, website recommendation~\cite{nilashi2016recommendation} and paper classification~\cite{citation}. Graph representation learning plays a crucial role in these tasks, focusing on acquiring general knowledge from abundant unlabeled graph data. Recent research has explored pre-training models on such data and applying them to downstream tasks to save training time and enhance performance~\cite{wu2020unsupervised,gcc,egi,grace,graphcl,rosa,zhu2023mario}. These efforts fall into two main categories.

The first group constructs training objectives based on domain-specific attributes and emphasizes pre-training and deployment on attributed graphs from the same domain~\cite{dgi,mvgrl,infograph}. That is, this approach requires consistent semantic meaning and feature dimensions across datasets, making it unsuitable for domain transfer.
For instance, DGI~\cite{dgi} and MVGRL~\cite{mvgrl} are traditional self-supervised learning frameworks tailored for specific attributed graphs~\cite{cui2020adaptive}. They are pre-trained and deployed on the same graphs.
However, using these models on different attributed graphs is not feasible due to inconsistent dimensions. For example, deploying a PubMed-pretrained model on the Cora dataset~\cite{sen2008collective} is unfulfillable, despite both scientific citation networks.

The second group focuses on learning transferable patterns through local structural information, enabling effective application to out-of-domain graph domains. This approach disregards node attributes during pre-training to avoid mismatches and facilitates the deployment of pre-trained models on diverse downstream datasets without relying on specific node attributes~\cite{gcc,egi}.
For instance, GCC~\cite{gcc} is a graph self-supervised pre-training framework designed to capture universal topological properties across multiple graphs by using structural information as node attributes. However, during deployment, this approach does not effectively utilize downstream informative node attributes. 
In scenarios where nodes represent papers and contain essential information like abstracts, neglecting these attributes can impact tasks like node classification.

Nonetheless, these approaches both encounter ``\emph{transferability-specificity dilemma}'':

\emph{transferability} $\times$ --- \emph{specificity} \checkmark: The first group pre-trains models using domain-specific features and deploys them on the same graph, but fails to achieve domain transfer.

\emph{transferability} \checkmark --- \emph{specificity} $\times$: The second group aligns the feature space with structural information to achieve domain transfer, but can not effectively utilize valuable downstream node attributes.

To overcome these challenges, we propose an innovative module for effective adaptation of pre-trained models to downstream datasets, compatible with existing pre-trained models. Specifically, we utilize universal structural pre-trained models~\cite{gcc} and incorporate unique features of downstream data as input conditions. Drawing inspiration from ControlNet~\cite{controlnet}, we feed structural information into the frozen pre-trained model and well-designed conditions into the trainable copy. The components are linked through zero MLPs, gradually expanding parameters from zero to incorporate valuable downstream attributes and safeguard against detrimental noise during finetuning.
To ensure that the pre-trained model (trainable copy) comprehends the condition effectively, we pre-process the condition input in a manner consistent with the pre-training strategy through our condition generation module. In essence, this approach enables us to utilize the specific statistics of downstream data, leading to more effective fine-tuning or prompt tuning (\emph{transferability} \checkmark --- \emph{specificity} \checkmark). This innovation opens the door to more effective and efficient deployment of pre-trained models in real-world Web applications. Through extensive experiments\footnote{In this study, our focus lies on node-level downstream tasks, excluding graph classification. The alignment of node (atom) attributes in molecules (one classical data type of graph classification) mitigates the challenges in graph transfer learning.}, we observe that our method can enhance the adaptability of pre-trained models on downstream datasets, achieving 2-3x performance gain on Cora\_ML and Amazon-Photo datasets. Furthermore, it surpasses training-from-scratch methods over 5\% absolute improvement on some datasets.

Our contributions can be concluded as:
\begin{itemize}
    \item We propose a novel deployment module coined as \ours to address the ``\emph{transferability-specificity dilemma}'' in graph transfer learning.
    \item We design a condition generation module to preprocess downstream-specific information into the pre-training data format, enabling the pre-trained model to understand the condition input effectively.
    \item Extensive experiments show that the proposed module significantly enhances the adaptability of pre-trained models on downstream datasets and can be seamlessly integrated with existing pre-trained models.
\end{itemize}

\section{Related Work}
\subsection{Graph Pre-training}
Graph pre-training aims to acquire a generalized feature extractor using abundant graph data. Existing self-supervised graph pre-training methods can be categorized as generative, contrastive, and predictive methods~\cite{wu2020unsupervised}. Generative methods like GAE~\cite{gae}, GraphMAE~\cite{gmae} focus on learning local relations by reconstructing features or edges. Contrastive methods~\cite{grace,graphcl,rosa} bring positive pairs closer and push negative pairs apart to learn global relations. Predictive methods involve creating pretext tasks manually based on data statistics to acquire generic knowledge.

In this research, our focus lies in Graph Contrastive Learning (GCL) methods, owing to their popularity and remarkable achievements~\cite{dgi,mvgrl,grace,graphcl,rosa,infograph,you2021graph,byol,zhu2023mario,zhu2023sgl}. The strategies employed by GCL methods can be broadly categorized into two primary groups. The first group formulates the training objective based on domain-specific features, such as DGI~\cite{dgi} and MVGRL~\cite{mvgrl}. However, these approaches inherently limit the transferability of models to other application domains, lacking versatility for effective application to attributed graphs from diverse domains. Contrastingly, the second group~\cite{gcc,graphcl} directs its attention towards learning transferable patterns by discerning local graph structures, thus completely circumventing the challenge of potentially unshared attributes. Nevertheless, real-world downstream datasets are often imbued with semantic attributes. Effectively leveraging this downstream-specific information within this paradigm remains an unresolved challenge.

\subsection{Graph Transfer Learning}
Graph transfer learning~\cite{lee2017transfer,transfer_survey,zhuang2020comprehensive} involves adapting pre-trained model on target data to economize training time and enhance downstream performance. Various strategies, such as domain adaptation~\cite{you2019universal}, multi-task learning~\cite{multitask}, and fine-tuning~\cite{sun2019fine}, can be employed to achieve transfer learning. 

In light of the remarkable achievements in pre-training techniques, this study focuses on fine-tuning. Initially, a generic model undergoes pre-training on extensive unlabeled graph data (source data). Subsequently, these pre-trained models are adapted for specific downstream tasks (target data). Notably, current fine-tuning methods~\cite{gcc} predominantly focus on adjusting pre-trained model parameters while simply incorporating target data.
However, a substantial challenge arises when the feature distribution of the target data diverges from that of the source data, potentially extending to differences in feature space. For example, the pre-trained model may have a fixed input dimension (\eg, 32) for structural attributes, whereas semantic attributes (\eg, keywords, abstract in paper) in the target data can vary across arbitrary dimensions. Traditional fine-tuning methods inadequately tackle this issue.

\sloppy
To address the non-trivial problem, we propose a deployment module, coined as \our, designed to incorporate downstream-specific information as input conditions. The condition will be processed to align with the format of pre-training data, enabling comprehension by pre-trained models, and steering the pre-trained model to predict more accurately, significantly enhancing the effectiveness of graph domain transfer learning.

\section{Background and Problem Formulation}
In this section, we will start with the notations we use throughout the rest of the paper in Sec.~\ref{sec:notations}. Subsequently, we will outline the specific problems under consideration in Sec.~\ref{sec:problem}.
\subsection{Notations}\label{sec:notations}
Let $\mathcal{G},\mathcal{Y}$ represent input and label space. $f_\phi(\cdot)$ represents graph predictor which consists of a GNN encoder $g_\theta(\cdot)$ and a classifier $p_\omega(\cdot)$. The graph predictor $f_\phi: \mathcal{G}\mapsto\mathcal{Y}$ maps instance $G=({A},{X}, {Y}) \in \mathcal{G}$ to label ${Y} \in \mathcal{Y}$ where ${A} \in \mathbb{R}^{N \times N}$ is the adjacency matrix and ${X} \in \mathbb{R}^{N \times d}$ is the node attribute matrix. Here, $N$, $d$ denote the number of nodes and attributes, respectively. Let $G_i$ denote a subgraph centered around node $i$ sampled from the original graph $G$.

\subsection{Problem Definition}\label{sec:problem}
\subsubsection{Universal Graph Representation Learning (UGRL)}
UGRL seeks a universal feature extractor, $g_\theta$, from unlabeled graph data for versatile application across diverse datasets. Challenges in node attribute-based pre-training, arising from varying and occasionally absent attributes, are addressed with our introduced structure pre-training models, collectively termed UGRL. These models efficiently transfer knowledge despite disparities in node attributes.

GCC\cite{gcc} is a classical structural pre-training method that leverages structural information as input. This approach aligns the input space across all datasets using structural information, facilitating domain transfer. To learn common knowledge, GCC will sample subgraphs $\{G_i\}_{i=1}^N$ from the original large graph $G$ and embed subgraphs with similar local structures closely through subgraph instance discrimination. Inspired by GCC, we employ generalized positional embedding as input features during pre-training. Formally, given an adjacency matrix $A$ and the corresponding degree matrix $D$, we conduct eigen-decomposition on its normalized graph Laplacian $U\Lambda U^T=I-D^{-\frac{1}{2}}AD^{-\frac{1}{2}}$. The top eigenvectors in $U$ will serve as generalized positional embedding. The GNN encoder $g_\theta:\mathbb{R}^{N\times k}\mapsto \mathbb{R}^{N\times l}$ maps the positional embedding ${P}\in\mathbb{R}^{N\times k}$ ($k$ set as 32 in this paper) to node embedding ${H}\in\mathbb{R}^{N\times l}$. To learn transferable structural patterns from positional embedding, we will maximize the mutual information between two similar subgraphs. Taking the InfoNCE loss\cite{cpc} as an example, the formulation follows:
\begin{equation}
\begin{aligned}
\mathcal{L}_{\text{MI}} & \left(g_\theta;G\right)=-\underset{G_i,G_i^\prime \in G}{\mathbb{E}}\left\|g_\theta(G_i)-g_\theta\left(G_i^\prime\right)\right\|^2 \\
+ & \underset{G_i \in G}{\mathbb{E}} \log \underset{G_j \in G}{\mathbb{E}}\left[  e^{\left\|g_\theta\left(G_i\right)-g_\theta\left(G_j\right)\right\|^2}\right],
\end{aligned}
\label{equ:con}
\end{equation}
where $g_\theta$ denotes GNN encoder with readout function, $G_i, G_i^\prime$ represents subgraphs centered around node $i$ sampled from the original graph $G$. The sampling strategy is random walk with restart which is also adopted in GCC~\cite{gcc} and RoSA~\cite{rosa}. So this method is scalable on large graphs. $G_i$ and $G_i^\prime$ share similar local structural information, serving as positive pairs, while $G_i$ and $G_j$ (sampled from different central nodes) act as negative samples. Through this self-supervised objective, UGRL obtains pre-trained models applicable to various downstream datasets, addressing specific tasks like node classification. This learning framework is commonly referred to as graph transfer learning.

\subsubsection{Graph Transfer Learning}
Graph transfer learning aims to leverage the universal knowledge within a pre-trained model, trained on source data, and apply it to target data for specific tasks.
There exists source data $\mathcal{D}^{\text{source}}$ and target data $\mathcal{D}^{\text{target}}$ from similar domains. In this paper, we assume source data is abundant but without labels and target data is limited but with labels. UGRL acquires pre-trained models $g_\theta^\star$ on the source data, which are then fine-tuned on the limited target data to accomplish specific tasks.

Take the downstream node classification tasks as an example, it involves learning a conditional probability $P({Y}\mid G;\phi)$ to categorize unlabeled nodes. To model this probability, the graph predictor $f_\phi(\cdot)=p_\omega\circ g_\theta^\star(\cdot)$ is employed where $p_\omega$ is a classifier and $g_\theta^\star$ is pre-trained GNN encoder obtained by the last part. The graph predictor is then optimized with training nodes from target data: 
\begin{equation}
    f_\phi^*=\argmin_{\phi} \mathbb{E}_{G\sim \mathcal{D}^{\text{target}}}\ell\left(f_\phi\left(X_{\text{train}},A\right),Y_{\text{train}}\right),
\end{equation}
where $X_{\text{train}}$ represents the attribute set of training nodes and $\mathbf{Y}_{\text{train}}$ denotes their labels, and $\ell(\cdot,\cdot)$ is cross-entropy loss. Finally, the optimal graph predictor $f_\phi^*$ is utilized for classifying testing nodes.

However, during pre-training, we only utilize structural information to obtain transferable pre-trained models while disregarding non-transferable node attributes. Sometimes, downstream data includes specific node attributes (\eg, age, gender, and interests) crucial for the task but incompatible with the pre-trained model (due to disparities in feature space and dimensions).
Incorporating these meaningful attributes into the model and guiding it towards superior performance represents a substantial challenge. We will present the solution to this challenge in Sec.~\ref{sec:method}.
\begin{figure}[htp]
    \centering
    \includegraphics[scale=0.6]{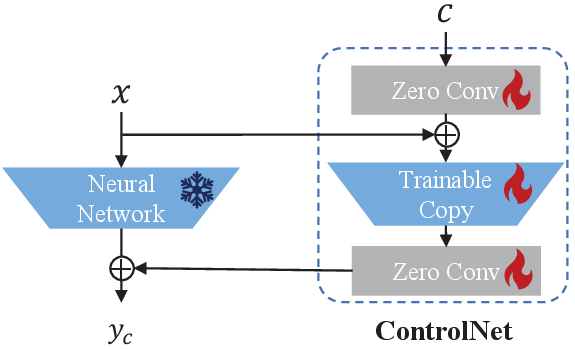}
    \caption{ControlNet injects conditions into neural network. $x$ represents original input and $c$ denotes condition input.}
    \label{fig:controlnet}
\end{figure}
\begin{figure*}
    \centering
    \includegraphics[scale=0.54]{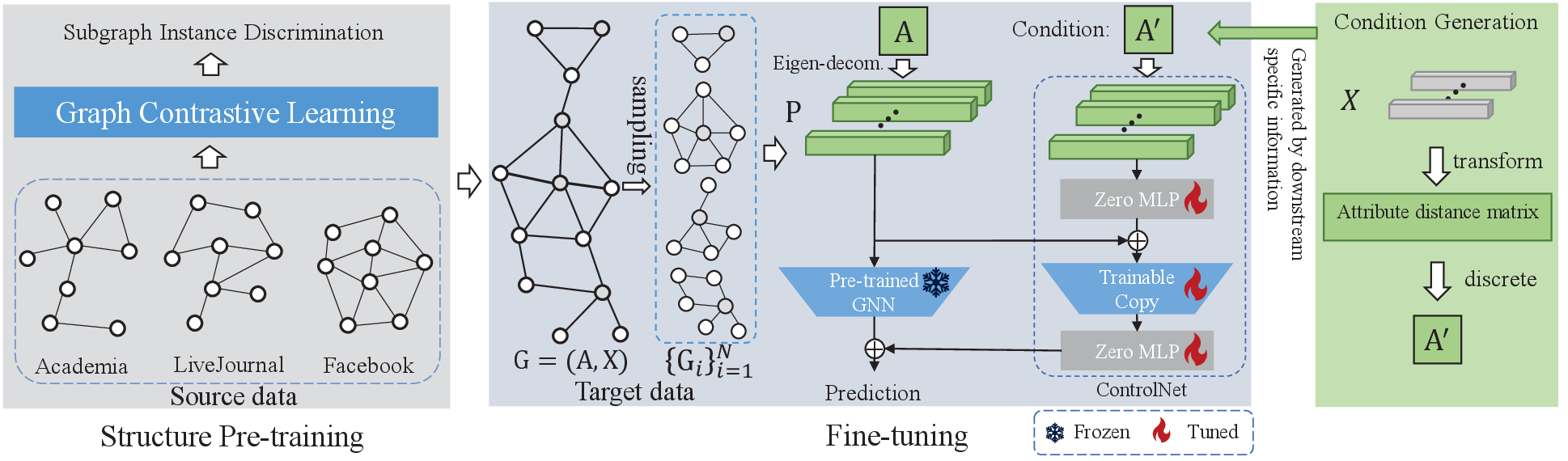}
    \caption{The pipeline of Graph Domain Transfer Learning with \our: Universal structure pre-training will be applied on extensive source data, then the pre-trained model will be deployed on target data with \ours, which includes Condition Generation and modified ControlNet. }
    \label{fig:framework}
\end{figure*}
\section{\our: Graph Transfer Learning with Conditional Control}\label{sec:method}
In this section, we will outline our approach to address the aforementioned challenge. Firstly, we retrospect the ControlNet in Sec.~\ref{sec:controlnet}. Then in Sec.~\ref{sec:condition}, we introduce the condition generation mechanism, which serves as the central component in \our. Subsequently, we detail our module \ours designed to adapt the pre-trained model to the target data in Sec.~\ref{sec:gcontrol}. Lastly, we demonstrate how to incorporate our module with fine-tuning and prompt-tuning techniques in Sec.~\ref{sec:finetune} and Sec.~\ref{sec:prompt}. Furthermore, we add complexity analysis in Appendix~\ref{app:time} due to the space limit.

\subsection{ControlNet}\label{sec:controlnet}
Firstly, let us retrospect the concepts of ControlNet~\cite{controlnet}. ControlNet is a neural network architecture designed to incorporate spatial conditioning controls into pre-trained diffusion models~\cite{kingma2021variational,stable_diff,croitoru2023diffusion} to generate customized images. Specifically, it freezes the pre-trained model and reuses the deep and robust encoding layers as a robust backbone (trainable copy) for acquiring diverse conditional controls. The trainable copy and original model are linked by zero convolution layers, progressively growing parameters from zero, ensuring a noise-free fine-tuning process~\cite{controlnet}. This approach allows us to control diffusion models with specified conditions. 

\subsection{Condition Generation}\label{sec:condition}
In graph domain, the downstream-specific node attributes pose compatibility challenges with the pre-trained model, primarily due to the disparities in terms of feature semantics and dimensions. One straightforward approach is to train a dedicated feature extractor for these attributes and integrate it with pre-trained models for prediction. However, this solution encounters three main issues:
(i) The downstream data is of a small scale, akin to few-shot scenarios~\cite{garcia2017few}, making training-from-scratch susceptible to overfitting on the training data and poor generalization on testing data.
(ii) The pre-trained model's assistance remains limited, failing to fully leverage its potential.
(iii) Selecting an appropriate feature extractor for node attributes is an open question, as different datasets may require different extractors. A brute-force approach trains with all choices and selects the best, incurring high training costs~\cite{you2020design}.

Our purpose is to enable the existing structural knowledge pre-training framework to utilize the node attributes of different downstream datasets in the fine-tuning or prompt-tuning stage. To achieve this, we draw inspiration from ControlNet~\cite{controlnet}, a neural network architecture that incorporates conditioning controls into large pre-trained text-to-image diffusion models. 
Considering the unique structure of the graph data, there are two primary distinctions that set our work apart from ControlNet:

1) \textbf{Motivation:} ControlNet aims to generate customized images through user instructions. However, in this study, we address the challenge of graph domain transfer learning by incorporating elaborate conditional control, generated from downstream specific information, into universal pre-trained models to solve "transferability-specificity dilemma".

2) \textbf{Technical details:} In ControlNet, the input condition for pre-trained text-to-image models is easily designed using sketches (\eg, cartoon line drawings, shape normals). In our study, we for the first identify that the downstream-specific information can be used as a comprehensible condition for graph transfer learning. Based on that, we propose a novel condition generation module for aligning the input space, which is of great significance and novelty.

The proposed condition generation module is depicted in Figure~\ref{fig:framework}. It utilizes the downstream-specific characteristics like node attributes to design the condition in a similar format to the adjacent matrix. 
Specifically, firstly, we measure the distance between nodes through a kernel function $\kappa(\cdot,\cdot)$. Thus we have a kernel matrix~\cite{kernelmatrix} (attribute distance matrix) ${K}\in\mathbb{R}^{N\times N}$ where ${K}_{ij}=\kappa(x_i, x_j)$. In this work, we use the linear kernel with normalized term (cosine similarity function) for computational simplicity:
\begin{equation}
    \kappa(x_i,x_j)=\frac{x_i^Tx_j}{\|x_i\|\|x_j\|}.
\end{equation}

Next, we discretize the kernel matrix using a threshold filter, where values greater than the threshold $v$ become $1$, and others become $0$:
\begin{equation}
{A}^\prime_{i,j} = 
\left\{
             \begin{array}{ll}
             0,\quad &\text{if}~ K_{i,j} \leq v \\
             \\
             1,\quad &\text{o.w.}
             \end{array}
\label{eq7}
\right.
\end{equation}
We call $A^\prime$ as feature adjacent matrix that aligns and maps node features of different dimensions and different semantics to the adjacency matrix space. Finally, we will use the same process during pre-training to obtain the generalized positional embedding $P^\prime$, which will be used during fine-tuning. For non-attributed target graphs, node embedding strategies like Node2Vec~\cite{node2vec} can be applied to generate node attributes. These attributes can then be utilized for creating conditions via our condition generation module.

Next, we elucidate the integration of ControlNet into graph domain, leveraging our designed condition to facilitate graph domain transfer learning.

\subsection{\our: Transfer Learning with ControlNet}\label{sec:gcontrol}
In this work, we draw inspiration from ControlNet to solve the challenges of graph domain transfer learning.
Considering the non-euclidean nature and oversmoothing problem~\cite{cai2020note} in graph domain, we substitute zero convolution layers with zero MLPs rather than zero graph convolution layers. We leverage universal structural pre-trained models and incorporate the downstream-specific information as condition input, effectively tackling the ``transferability-specificity dilemma."
The structural information of target data will be fed to the frozen pre-trained model (to avoid catastrophic forgetting~\cite{liu2021overcoming,kirkpatrick2017overcoming,ninini,ninini2}) and the elaborate condition (generated by Condition Generation Module) will be fed into the trainable copy. 
These two components are linked by zero MLPs, gradually growing the parameters from zero. This approach ensures that no harmful noise affects the fine-tuning process while progressively incorporating downstream-specific information.

The procedure of our method can be formalized as follows:
\begin{equation}
    {H}_c=g_\theta^\star(P)+\mathcal{Z}_2(g_c(P+\mathcal{Z}_1(P^\prime))),
\end{equation}
where $\mathcal{Z}_1$ and $\mathcal{Z}_2$ represent the first and the second zero MLP, and $g_c(\cdot)$ represents the trainable copy of the pre-trained encoder $g_\theta^\star$. Similar to ControlNet, because the parameters of the zero MLP are set to 0 during initialization, we have $\mathcal{Z}_2(g_c(P+\mathcal{Z}_1(P_c))) = 0$. Hence, during initialization, our model's output aligns with using the pre-trained encoder alone. Throughout optimization, the downstream-specific information is progressively integrated.

\subsection{Fine-tuning with \our}\label{sec:finetune}
Given an input graph \( G = ({A},{X}) \), our process begins with preprocessing the graph data, involving subgraph sampling, generalized positional embedding calculation, and condition generation. In the training phase, we keep the parameters of the pre-trained GNN encoder \( g_\theta \) fixed to prevent catastrophic forgetting. The original positional embedding \( P \) is input into \( g_\theta^\star \), and the generated condition is input into the ControlNet. The resulting representations \( H_c \) are utilized for specific tasks. For example, in the node classification task, a linear classifier is added to map these representations to predicted labels. The classification error is then calculated using the cross-entropy loss function. All components are optimized in an end-to-end manner. The entire procedure is outlined in Algorithm~\ref{alg:GControl}.

\subsection{Graph Prompt Tuning with \our}\label{sec:prompt}
\begin{figure}
    \centering
    \includegraphics[scale=0.55]{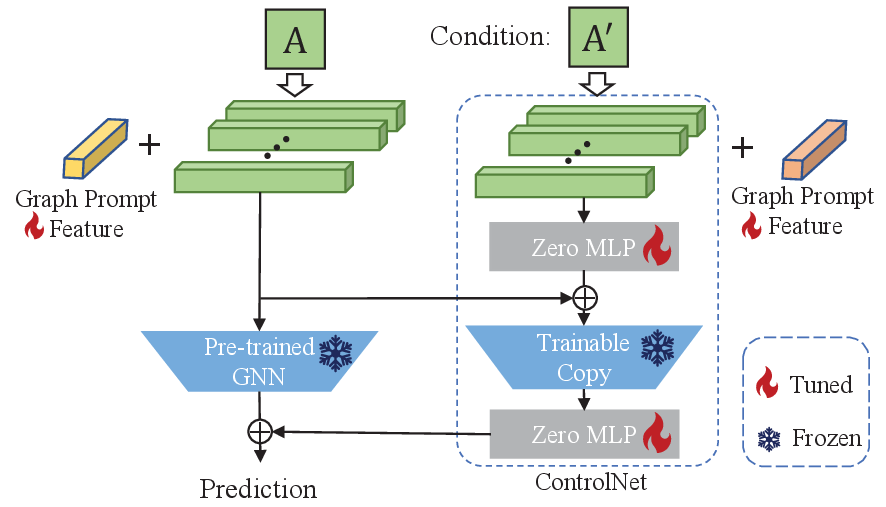}
    \caption{Graph prompt tuning with \our.}
    \label{fig:prompt}
\end{figure}

In the last section, we introduce how to perform fine-tuning with our method on target data.
In scenarios where training data for the target dataset is notably scarce (\eg, fewshot setting), tuning all parameters can result in overfitting and difficulties in generalizing effectively on the test set. To address these challenges, graph prompt tuning methods~\cite{gppt,gpf,zhu2023sgl,graphprompt,prog} have emerged which focus on tuning only a few parameters of the prompt. In this section, we will demonstrate that our method (\our) can seamlessly integrate with existing graph prompt methods, significantly enhancing downstream performance. Taking the GPF graph prompt tuning method~\cite{gpf} as an example, the workflow is illustrated in Figure~\ref{fig:prompt}. Firstly, two trainable graph prompt features $p,p^\prime \in \mathbb{R}^{1\times k}$ are randomly initialized. Then, these prompt features are broadcasted to be added to the input features. The formulation is as follows:

\begin{equation}
{H}_c=g_\theta^\star(P+q)+\mathcal{Z}_2(g_c((P+q)+\mathcal{Z}_1(P^\prime + q^\prime))),
\end{equation}
In contrast to the previous section, in graph prompt tuning, the parameters of the trainable copy \( g_c \) will be frozen to prevent overfitting. Besides, more intricate graph prompt methods, such as All-in-One (ProG)~\cite{prog}, can be integrated with our method to enhance downstream performance. Detailed experiments are in Sec.~\ref{sec:fewshot}.

\section{Experiments}
In this section, we first introduce the datasets, baselines, and experimental setup in Sec.~\ref{sec:dataset},~\ref{sec:baseline} and \ref{sec:setup} respectively.  
Secondly, we conduct main experiments under fine-tuning (Sec.~\ref{sec:transfer_exp}) and prompt tuning (Sec.~\ref{sec:fewshot}) to prove the effectiveness of \our.
We then perform an ablation study to demonstrate the effectiveness of each proposed component in Sec.~\ref{sec:ablation}. Lastly, we analyze the convergence of \ours (Sec.\ref{sec:convergence}) and the impact of important hyper-parameters (Sec.~\ref{sec:sensitivity}). 
\subsection{Datasets}\label{sec:dataset}
\textbf{Pre-training datasets.}\quad 
The details of the pre-training datasets are presented in Table~\ref{tab:pretraining_data} and Appendix~\ref{app:datasets}. Notably, these datasets are substantial in scale, with the largest graph comprising approximately 4.8 million nodes and 85 million edges.

\noindent\textbf{Downstream datasets.}\quad
We select eight public benchmark datasets as target data that include four attributed datasets (\ie, Cora\_ML, Amazon-Photo, DBLP, and Coauthor-Physics), and four non-attributed datasets (\ie, USA-Airport, Europe-Airport, Brazil-Airport, and H-index) to evaluate the effectiveness of \our. The statistics of datasets is in Table~\ref{tab:dataset}. Detailed illustrations of these datasets can be found in Appendix~\ref{app:datasets}.
\begin{table}[htp]
    \centering
    \caption{Statistics of datasets. These datasets can be further classified into attributed graphs and non-attributed graphs.}
    \scalebox{0.95}{
    \begin{tabular}{c|cccc}
    \toprule 
                         & \#Nodes & \#Edges & \#Attributes & \#Classes \\ \midrule
        Cora\_ML         & 2,995   & 16,316  & 2,879        & 7         \\
        Amazon-Photo     & 7,650   & 238,162 & 745          & 8         \\
        DBLP             & 17,716  & 105,734 & 1,639        & 4         \\
        Coauthor-Physics & 34,493  & 495,924 & 8,415        & 5         \\ \midrule
        USA-Airport      & 1,190   & 27,198  & -            & 4         \\
        Europe-Airport   & 399     & 5,995   & -            & 4         \\
        Brazil-Airport   & 131     & 1,047   & -            & 4         \\
        H-index          & 5,000   & 44,020  & -            & 44        \\
    \bottomrule    
    \end{tabular}}
    \label{tab:dataset}
\end{table}
\subsection{Baselines}\label{sec:baseline}
We evaluate \ours with five self-supervised pre-training methods (using GIN as encoder): \textbf{GCC}~\cite{gcc}, \textbf{GRACE}~\cite{grace}, \textbf{simGRACE}~\cite{xia2022simgrace}, \textbf{COSTA}~\cite{costa}, and \textbf{RoSA}~\cite{rosa}.
Detailed descriptions of these methods are available in Appendix~\ref{app:baseline}. Except for GCC, other pre-training methods are designed for attributed graphs. To integrate them into our setting, we replace their input with structural information, disregarding the original node attributes during pre-training.
To demonstrate the superiority of our approach over training from scratch, we compare it with the supervised GIN model, initialized randomly and trained on target data. Notably, GCC's encoder is based on GIN but with a little different implementation (\eg, incorporates additional information as input). GCC (rand) signifies the utilization of a randomly initialized GCC encoder, trained from scratch on the target data. Additionally, we include two baselines that only use node attributes (\ie, MLP~\cite{mlp}) and structural information (\ie, Node2Vec~\cite{node2vec}) of downstream datasets to demonstrate the effectiveness of both for classification. 
Considering the abundance of source data and its occasional unavailability (we only have access to pre-trained models), domain adaptation baselines~\cite{you2019universal} are not included in this work.

\begin{table*}[htp]
\caption{Experimental results of baselines and our method on downstream datasets. In the data column, $A$ represents adjacent matrix, $X$ denotes node attribute matrix and $X_{PE}$ means positional embeddings. Rows with gray background denote our method.}
\vspace{-1em}
    \centering
    \begin{tabular}{ll|cccc|cccc}
    \toprule
Data        &Methods                 & Cora\_ML   & Photo      & DBLP        & Physics     & USA        & Europe     & Brazil     & H-index     \\ \midrule
X        & MLP\cite{mlp}                       & 60.31±2.96 & 77.56±2.42 & 64.47±1.36 & 88.90±1.10  & -& -& -&-\\
A        &Node2Vec\cite{node2vec}                   & 69.93±1.27 & 84.08±0.63 & 77.52±0.38 & 88.13±0.39 & 59.59±2.04 & 47.92±3.66 & 46.53±8.41 & 75.02±0.50\\
A,$\text{X}_{PE}$&GIN\cite{gin}                & 29.94±1.37 & 30.41±1.07 & 57.53±0.78  &              54.76±0.69&            56.33±1.90&            49.72±3.05&             57.63±8.96& 69.90±1.26\\ 
A,X      &GIN\cite{gin}                        & 69.57±3.65 & 79.71±4.72 & 74.62±3.00 & 92.02±2.79 & 58.89±2.70& 47.85±4.86& 58.52±9.98&72.23±1.20\\
\midrule
A,$\text{X}_{PE}$&GCC (rand)          & 26.34±1.40 & 26.15±1.20 & 53.46±0.79  & 54.30±0.68 & 54.85±2.31 & 42.60±3.31& 51.20±8.49& 64.18±1.83\\
A,$\text{X}_{PE}$&$\text{GCC}$\cite{gcc}       & 31.14±1.23 & 35.51±0.54 & 57.02±0.68  & 56.25±0.37& 55.80±2.23 & 47.35±3.44& 57.92±9.00 & 70.31±1.89\\
\rowcolor{Gray}
A,$\text{X}_{PE}$,X &\quad+\our      & 77.43±1.62 & 88.65±0.60 & 80.25±0.90 & 94.31±0.12  & 57.03±2.21& 50.53±3.43 & 59.28±8.14& 73.55±0.70\\  \midrule
A,$\text{X}_{PE}$        &GRACE\cite{grace}      & 30.74±1.48 & 32.64±1.57 & 58.43±0.37 & 59.86±1.96 & 5768±1.75 & 50.49±2.90  &       57.98±9.45&69.68±2.18\\
\rowcolor{Gray}
A,$\text{X}_{PE}$,X    &\quad+\our   & 77.26±1.50 & 88.78±0.61 & 80.42±0.65 & 94.12±0.24 & 58.94±1.84 & 52.83±3.10&59.92±7.59&74.47±0.07\\ \midrule
A,$\text{X}_{PE}$ &simGRACE\cite{xia2022simgrace}          & 30.39±1.82 & 33.62±1.52 & 57.87±0.32 & 59.82±2.93 & 57.11±1.90 & 50.22±3.91& 58.09±8.50&69.65±1.50\\
\rowcolor{Gray}
A,$\text{X}_{PE}$,X  &\quad+\our     & 77.34±1.08 & 89.66±0.56 & 80.33±0.69 & 94.03±0.47& 59.40±1.62 & 51.15±3.17& 59.41±7.66&76.10±0.70\\ \midrule
A,$\text{X}_{PE}$        &RoSA\cite{rosa}       & 30.96±0.81 & 33.42±1.59 & 56.41±0.70&  60.14±2.48& 57.18±2.02 &50.32±3.78 & 58.99±8.30&69.80±2.48\\
\rowcolor{Gray}
A,$\text{X}_{PE}$,X    &\quad+\our   & 77.40±1.06 & 89.35±0.61 & 80.23±0.79 & 94.22±0.26 & 58.71±1.35 &51.89±2.69& 59.16±7.13&74.22±1.46\\ \midrule
A,$\text{X}_{PE}$        &COSTA\cite{costa}      & 30.07±1.31 & 33.22±1.28 & 59.01±0.19 & 59.96±3.29 & 57.07±2.53 &50.33±3.64& 59.55±9.30&68.49±2.10\\
\rowcolor{Gray}
A,$\text{X}_{PE}$,X    &\quad+\our   & 76.63±1.67 & 89.17±1.14 & 80.74±0.65 & 94.02±0.31 & 59.00±1.82 &51.88±3.08& 62.16±6.95&73.57±2.17\\
    
    \bottomrule
    \end{tabular}
    \label{tab:transfer}
    \vspace{-0.5em}
\end{table*}

\subsection{Transfer Learning (Standard Fine-tuning)}\label{sec:transfer_exp}
In this section, we will conduct experiments with standard fine-tuning on target data to evaluate our method.
\subsubsection{Experimental Setup}\label{sec:setup}
For pre-trained models, GCC~\cite{gcc} is pre-trained on abundant unlabeled large graphs (\eg, Facebook~\cite{ritchie2016scalable}, LiveJournal~\cite{backstrom2006group}), we use their released pre-trained checkpoint\footnote{https://github.com/THUDM/GCC}. In the case of GRACE, simGRACE, RoSA, and COSTA, we perform pre-training on the downstream graphs, excluding node attributes. During fine-tuning, we incorporate the original node attributes. All pre-training methods use a 4-layer Graph Isomorphism Network (GIN)~\cite{gin} with 64 hidden units as encoders.

Regarding data splitting, we randomly divide the training and testing data into a 1:9 ratio, and the results represent the mean accuracy with a standard deviation computed over 20 independent experiments. Additional details and hyperparameters can be found in Appendix~\ref{app:hyper}.

\newcolumntype{g}{>{\columncolor{Gray}}c}
\newcommand{\CC}[1]{\cellcolor{white!#1}}
\begin{table}[htp]

\caption{Experimental results of fine-tuning (FT) and prompt tuning (PT) under few-shot settings (3-shot and 5-shot).}
\vspace{-1em}
    \centering
    \scalebox{0.85}{
    \begin{tabular}{c|g|gg|gg}
    \toprule
    \rowcolor{white}
            &       & \multicolumn{2}{c|}{USA}  & \multicolumn{2}{c}{Europe} \\ 
    \rowcolor{white}
            &       & 3-shot      & 5-shot      & 3-shot & 5-shot \\ \midrule
\rowcolor{white}
\multirow{2}{*}{FT}
&Finetuned GIN      & 34.28±4.06 & 35.73±4.45& 37.99±4.38 & 40.61±3.18     \\ 
&\cellcolor{white}Finetuned GCC      & \cellcolor{white}48.75±4.76 & \cellcolor{white}51.76±4.98& \cellcolor{white}45.08±4.24 & \cellcolor{white}48.62±3.77\\ 
\midrule
\rowcolor{white}
\multirow{4}{*}{PT}
&GCC+GPF\cite{gpf}  & 49.10±4.70 & 50.78±5.19& 47.10±3.57 & 49.69±3.47\\
&Ours+GPF\cite{gpf}  & 50.40±3.33 & 53.05±4.52& 47.50±3.99 & 50.29±2.42\\ \cmidrule{2-6}
\rowcolor{white}
&GCC+ProG\cite{prog}& 48.80±4.36 & 49.36±5.64& 46.04±4.38 & 48.47±3.67\\
&Ours+ProG\cite{prog}& 49.73±4.34 & 52.61±5.22& 46.81±4.39 & 50.65±2.93 \\
    \bottomrule
    \end{tabular}}
    \vspace{-1em}
    \label{tab:fewshot}
\end{table}

\subsubsection{Analysis}
From Table~\ref{tab:transfer}, we can draw the following conclusions: firstly, structural pre-training methods can learn transferable structural patterns because $\text{GCC}$ surpasses GCC(rand)\footnote{GCC(rand) refers to a randomly initialized encoder of GCC, trained from scratch on target data, focusing on structural information.} with comparable margins. For instance, on H-index and Cora\_ML datasets, GCC achieves over 5\% absolute improvement compared to GCC(rand).

\sloppy
Secondly, applying structural pre-training methods directly to target attributed graphs fails to achieve satisfactory performance and notably lags behind training-from-scratch methods (\eg, GIN(A,X)) on target data. This underscores the essential role of downstream-specific information (\eg, node attributes) for optimal performance. For instance, on the DBLP dataset, GCC achieves only 57\% accuracy, lagging behind GIN(A,X) by approximately 17\%.

Thirdly, deploying structural pre-trained models on target data with \ours significantly enhances performance. For instance, on Cora\_ML and Photo datasets, our method achieves 2-3x performance gains compared to direct deployment. Moreover, when pre-trained models are combined with \ours, intelligently leveraging downstream-specific information, they outperform training-from-scratch methods on target data, showcasing \ours' ability to fully harness the potential of pre-trained models. Even for non-attributed target data, our method can enhance downstream performance with additional node embeddings from Node2Vec~\cite{node2vec}. Specifically, GRACE with \our outperforms GIN(A,$\text{X}_{PE}$) by approximately 5\% absolute improvement.

These statistics show the effectiveness of our module for deploying universal pre-trained models on target data.

\begin{table*}[htp]
\caption{Ablation studies for \ours by masking each component. We use \textbf{boldface} and \underline{underlining} to denote the best and the second-best performance, respectively.}
\vspace{-0.5em}
    \centering
    \fontsize{8.5pt}{8.5pt}\selectfont
    \begin{tabular}{ll|cccc|cccc}
    \toprule
Data        &Methods                   & Cora\_ML   & Photo      & DBLP       & Physics     & USA        & Europe     & Brazil     & H-index     \\ \midrule

A,$\text{X}_{PE}$,X&Ours                & \textBF{77.43±1.62}& \textBF{88.65±0.60}& \textBF{80.25±0.90}& \textBF{94.31±0.12}& \textBF{57.03±2.21}& \textBF{50.53±3.43}& \textBF{59.28±8.14}&\textBF{73.55±0.70}\\
A,$\text{X}_{PE}$,X&Ours (soft C)        & 27.42±3.32& 51.85±7.30& 51.42±3.37& 77.75±2.51& 55.03±2.57& \underline{48.67±3.95}& \underline{58.81±8.31}&\underline{73.26±0.63}\\
A,$\text{X}_{PE}$,X&Ours (w/o zero)      & {71.93±2.73}& 81.30±4.33& \emph{77.24±5.55}& \underline{93.70±0.25}& {55.35±3.08}& {48.49±3.39}& {58.03±9.91}&71.71±1.63\\
A,$\text{X}_{PE}$  &Ours (w/o CG)        & 20.29±2.09& 27.31±3.09& 49.62±3.54& 51.96±0.70& 55.31±2.32& 47.93±3.41& 56.65±8.76&72.79±0.53\\
A,X         &Ours (w/o frozen pre.)      & \underline{75.06±1.90}& {82.38±4.72}& \underline{79.22±1.58}& {93.56±0.22}& \underline{55.85±2.71}& 47.75±3.04& 56.57±8.65&{73.16±0.72}\\
A,$\text{X}_{PE}$,X&Simple Cat.        & 66.64±1.41& \underline{82.80±0.95}& 67.32±0.60& 90.56±0.46& 53.59±4.74& 45.28±5.29& 57.67±7.14&72.74±1.16\\

    \bottomrule
    \end{tabular}
    \label{tab:ablation}
    
\end{table*}

\subsection{Few-shot Classification (Prompt Tuning)}\label{sec:fewshot}
In many real-world scenarios, the target data is notably limited, with only a few training samples for each class. Few-shot learning is a well-known case of low-resource scenarios. Standard fine-tuning tends to overfit on the training data, leading to poor generalization. To solve these problems, Graph prompt tuning emerged which can align the training objectives and train a few parameters of prompt. In this section, we will perform experiments of existing graph prompt tuning with \ours under few-shot setting.
\subsubsection{Baselines \& Experimental setup}
We choose two graph prompt methods, GPF~\cite{gpf} and ProG~\cite{prog}, because they are not limited to specific pre-trained GNN models. Other graph prompt methods like GPPT~\cite{gppt}, GraphPrompt~\cite{graphprompt}, and SGL-PT~\cite{zhu2023sgl} heavily rely on specific pre-trained models will not included in this study. 
GPF introduces trainable graph prompt features applied to the original graph, imitating any graph manipulations. ProG is a more complex version, inserting a prompt graph comprising multiple prompt features and relations into the original graph.
For the pre-trained model, we adopt GCC here for simplicity. 

`Finetuned GIN' and `Finetuned GCC' refer to randomly initialized GIN and pre-trained GCC fine-tuned on target data. `GCC+GPF' indicates pre-trained GCC prompt tuning on target data with GPF, while `Ours+GPF' involves deploying pre-trained GCC with \ours using GPF as prompt tuning. `GCC+PorG' and `Ours+PorG' use ProG as prompt tuning method.

Target data is split into 1:9 for candidate and testing data. In 3-shot(5-shot) setting, 3(5) samples per class are selected from candidate data for training. Results show mean accuracy with standard deviation over 20 experiments. More details in Appendix~\ref{app:hyper}.

\subsubsection{Analysis} From Table~\ref{tab:fewshot}, we can draw the following conclusions: firstly, we can see the `Finetuned GIN' achieves the worst performance because the training from scratch will easily overfit on limited training data. `Finetuned GCC' achieves a decent result but lags behind prompt methods with few training samples. For example, under 3-shot, `Finetuned GCC' is outperformed by `GCC+GPF' and `GCC+ProG', but surpasses them with more training data (5-shot). This highlights the greater effectiveness of prompt methods under limited resources.

Secondly, with prompt tuning, our method \ours can still enhance the downstream performance. Specifically, both `Ours+ProG' and `Ours+GPF' outperform their corresponding baselines (`GCC+ProG' and `GCC+GPF') by 2\% absolute improvement. In the small-scale dataset like 5-shot Europe-Airport, `Ours+ProG' reaches comparable performance to full-shot in the last section.

\subsection{Ablation Studies}\label{sec:ablation}
In this section, we assess the effectiveness of \ours by masking each component.
Ours (soft C) uses soft attribute distance matrix $K$ as condition. Ours (w/o zero) removes zero MLPs in ControlNet. Ours (w/o CG) removes the condition generation and ControlNet, similar to finetuing GCC. Ours (w/o frozen pre.) excludes the frozen pre-trained model branch, utilizing only the ControlNet branch. Lastly, `Simple Cat.' signifies a basic approach: training a dedicated feature extractor for downstream attributes from scratch and integrating it with pre-trained models for predictions. As for the pre-trained model, we use GCC in this section.

Each component is crucial for the effectiveness of the method, as shown in Table~\ref{tab:ablation}. Specifically, Ours (w/o CG) performs poorly on attributed datasets, emphasizing the importance of downstream-specific information. Ours (soft C) also underperforms, highlighting the significance of aligning the format of condition and input during pre-training. Ours (w/o zero) lags behind \ours by a comparable margin, indicating the importance of zero MLPs in linking the frozen pre-trained model and the trainable copy, avoiding detrimental noise during fine-tuning. Ours (w/o frozen pre.) is also inferior to \our, indicating the effectiveness of incorporating common knowledge from the pre-trained model. Finally, `Simply Cat.' achieves subpar results on most datasets, emphasizing the risks of overfitting when training from scratch on limited data, especially in smaller datasets like Cora\_ML (10\% lower than \our).

\subsection{Convergence Analysis}\label{sec:convergence}
Earlier sections demonstrate the efficacy of our method in performance enhancement. In this section, we delve into the convergence speed analysis of \our. Here, `GIN' denotes training from scratch, while \ours signifies using GCC as the pre-trained model and fine-tuning it with \our.

\begin{figure*}
    \centering
    \includegraphics[scale=0.78]{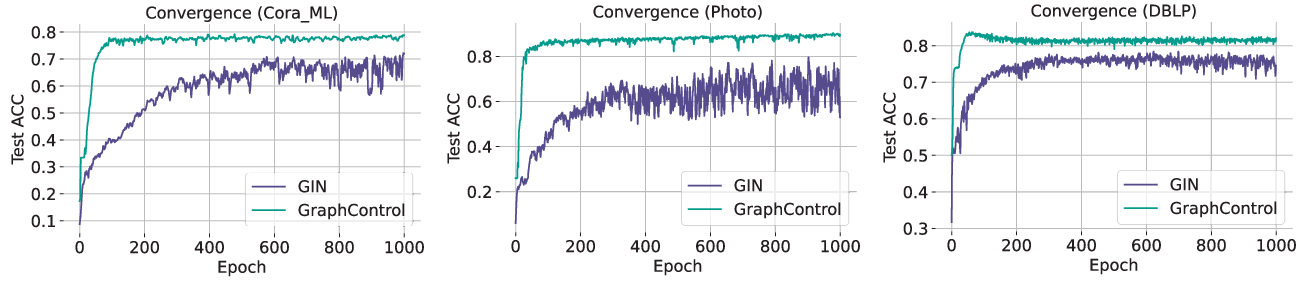}
    \caption{The convergence analysis on GIN and \our.}
    \vspace{-1em}
    \label{fig:convergence}
\end{figure*}

As depicted in Figure~\ref{fig:convergence}, \ours achieves convergence within 100 epochs on all datasets, whereas GIN reaches the best performance around 600 epochs on Cora\_ML and Photo datasets, exhibiting instability. Our approach not only improves performance but also notably reduces training time in downstream applications.

\subsection{Sensitivity Analysis}\label{sec:sensitivity}
In this section, we analyze crucial hyperparameters, starting with the impact of the threshold used in condition generation, followed by an analysis of the subsampling hyperparameters.
\subsubsection{Analysis on threshold for discretization}
\begin{figure}
    \centering
    \includegraphics[scale=0.35]{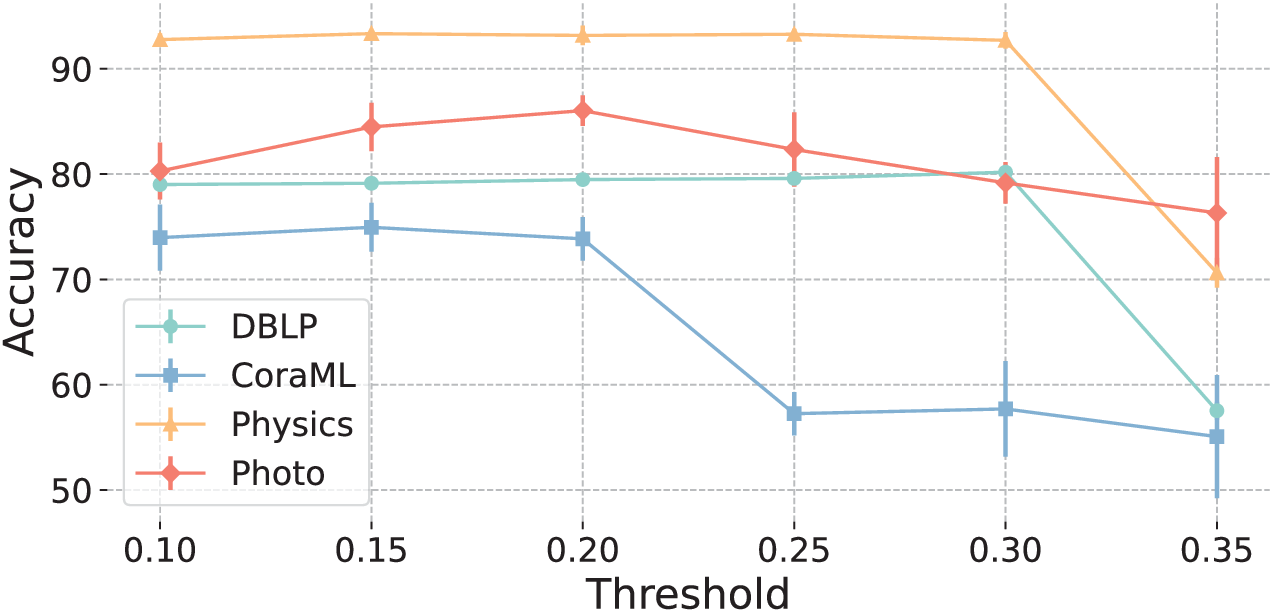}
    \caption{Sensitivity analysis on threshold.}
    \vspace{-1em}
    \label{fig:sensitivity}
\end{figure}

In the process of condition generation, we discretize the attribute distance matrix using a specific threshold, converting it into a feature-based adjacency matrix to align with the input space during pre-training. Empirically, we explore the impact of this threshold, ranging from 0.1 to 0.35. The results, shown in Figure~\ref{fig:sensitivity}, indicate stable performance from 0.1 to 0.2 on DBLP, Cora\_ML, and Physics datasets. However, when the threshold exceeds 0.3, most datasets experience a rapid drop in performance due to the matrix becoming overly sparse and providing limited information. 
Optimal thresholds range from 0.15 to 0.2, guiding our experiments across most datasets.

\subsubsection{Analysis on hyper-parameters of subsampling} 
\begin{figure}
    \centering
    \includegraphics[scale=0.485]{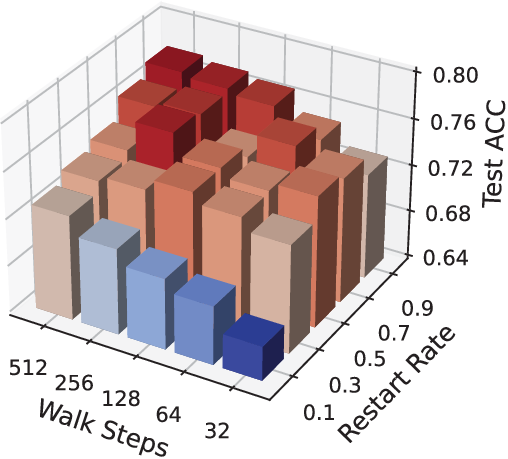}
    \includegraphics[scale=0.485]{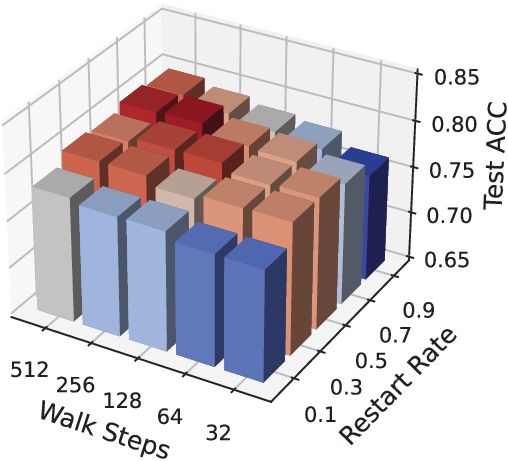}
    \caption{Analysis of subsampling hyperparameters on Cora\_ML (left) and DBLP (right) datasets.}
    \vspace{-1em}
    \label{fig:subsampling}
\end{figure}

In this work, random walk with restart serves as the subsampling technique, with walk steps and restart rate as pivotal hyperparameters. Walk steps are selected from $\{32, 64, 128, 256, 512\}$, and the restart rate spans $\{0.1, 0.3, 0.5, 0.7, 0.9\}$. Based on Figure~\ref{fig:subsampling}, optimal results are observed with 256 and 512 walk steps, alongside restart rates of 0.7 and 0.9. For memory efficiency, we standardize walk steps to 256 across all datasets and set the restart rate to 0.8 for most datasets.

\section{Conclusion}
In this work, we propose a novel deployment module coined as \ours to address the challenges of the `pre-training and finetuning (or prompt-tuning)' paradigm in graph domain transfer learning. 
\ours seamlessly integrates with existing universal structural pre-trained models, significantly boosting their performance on target data by intelligently incorporating downstream-specific information.
Specifically, to achieve this, we draw inspiration from ControlNet and apply its core concepts to graph domain transfer learning. 
Downstream-specific information is processed into conditions using our condition generation module and gradually integrated for enhanced performance.
Extensive experiments on diverse real-world datasets demonstrate the superiority of \ours in fine-tuning and prompt tuning scenarios, substantially improving the adaptability of pre-trained models on target data.
\section*{Acknowledgments}
This work was supported by the NSFC (No. 62272411), Key Research and Development Projects in Zhejiang Province (No. 2024C01106), the Tencent WeChat Rhino-Bird Special Research Program (Tencent WXG-FR-2023-10), and the National Key Research and Development Project of China (2018AAA0101900).


\bibliographystyle{ACM-Reference-Format}
\bibliography{refs}

\newpage
\appendix
\section{Algorithm}
The complete procedure of our method with fine-tuning is outlined in Algorithm~\ref{alg:GControl}. Given an input graph \(G=(A,X,Y)\), we employ a subsampling function $\mathcal{T}$ to sample subgraphs for each node. Subsequently, we generate the condition \(P_i^\prime\) using node attributes and positional embeddings \(P_i\) using adjacency matrix for each subgraph \(G_i\). The training dataloader $\mathcal{S}$ is then created with a batch size of 128 for subgraphs of training nodes. During each iteration, batched graphs are inputted into the frozen pre-trained model \(g_{\theta}^\star\), while the condition is fed into the trainable copy \(g_{\theta_c}\). These two components are interconnected using zero MLPs. Finally, the representations $H$ are passed through a classifier $p_\omega$, with the cross-entropy loss $\mathcal{L}$ utilized to compute the classification error $\ell_{\text{sup}}$. The parameters of the trainable copy, zero MLPs, and classifier are optimized by minimizing the loss.
\begin{algorithm}[tb]
    \caption{\ours algorithm}
    \label{alg:GControl}
    \KwIn{Frozen pre-trained GNN encoder $g_\theta^\star$, trainable copy of pre-trained model $g_{\theta_c}$, two zero MLPs $\mathcal{Z}_1, \mathcal{Z}_2$ with parameters $\theta_{Z_1}, \theta_{Z_2}$, random initialized linear classifier $p_\omega$, input graph $G=(A,X,Y)$, sampler function $\mathcal{T}(G,i)$, training epochs $E$, learning rate $\eta$.} 
    \KwOut{Optimized models, $g_{\theta_c}$, $\mathcal{Z}_1$, $\mathcal{Z}_2$, $p_\omega$}
    \tcc{subsampling}
    \For{$i\leftarrow 1$ \KwTo $N$}{
    $G_i = \mathcal{T}(G,i)=(A_i,X_i)$\;
    Generate condition $A_i^\prime$ through Condition Generation Module using node attributes\;
    Generate positional embedding ${P}_i$ from graph adjacency matrix ${A}_i$ and conditional positional embedding $P^\prime_i$ from attribute adjacency matrix $A_i^\prime$\;
    $G_i=(A_i,P_i, P^\prime_i,y_i)$
    }
    $\mathcal{S} = \{G_i\}, i=1,..., N$ \tcp*{collect training samples}
    \For{$e\leftarrow 1$ \KwTo $E$}{
    Sampled batch $\mathcal{B}=\{G_i\}_{i=1}^B \in \mathcal{S}$\;
    \tcc{For symbol unclutter, we omit subscript}
    Batched graph $G=(A,P,P^\prime,Y)$\;
    \tcc{Forward}
    ${H}\leftarrow g_\theta(P)+\mathcal{Z}_2(g_{\theta_c}({P}+\mathcal{Z}_1({P}^\prime)))$ \;
    $\ell_{\text{sup}}\leftarrow \mathcal{L}(p_\omega({H}), {Y})$ \;
    \tcc{Backward}
    $\theta_c \leftarrow \theta_c - \eta \nabla_{{\theta_c}} \ell_{\text{sup}}$; $\theta_{Z_1} \leftarrow \theta_{Z_1} - \eta \nabla_{{\theta_{Z_1}}} \ell_{\text{sup}}$; $\theta_{Z_2} \leftarrow \theta_{Z_2} - \eta \nabla_{{\theta_{Z_2}}} \ell_{\text{sup}}$ \;
    }
\end{algorithm}

\section{Complexity Analysis}\label{app:time}
The time and memory complexity of our method is comparable to GCC. Specifically, 'GCC+GraphControl' comprises two components: the frozen encoder and ControlNet (consisting of a trainable copy and two zero MLPs). During training, we freeze the pre-trained encoder and exclusively fine-tune the trainable copy and zero MLPs. Consequently, the time complexity of the backward pass and the memory complexity of our method are approximately equivalent to GCC. Formally, the analysis of time complexity and memory complexity follows as:

Given a sparse input graph $G=(A,X,X_{\text{PE}},Y)$, where $N$ denotes the number of nodes, $D$ represents the number of node attributes, $X \in \mathbb{R}^{N\times D}$ is the attribute matrix, and $X_{\text{PE}} \in \mathbb{R}^{N \times K}$ is the positional embedding with $(K=32 \ll D=745)$. Let $H$ represent the hidden size and $L$ denote the number of layers in the model. In some cases, $H$ is comparable to $D$, so we assume them to be the same in this analysis for simplicity. The parameters of the two zero MLPs are in $\mathbb{R}^{K\times K})$ and $(\mathbb{R}^{D\times D})$. The following results can be derived: $\mathcal{O}(ED)$ denotes message passing, i.e., $AX$, and $\mathcal{O}(ND^2)$ denotes feature transformation, \ie, $XW$, in each layer. For the two zero MLPs, matrix multiplication and summation amount to $NK+NK^2$ and $ND+ND^2$. Therefore, the time and space complexity follows:

\begin{table}[htp]
    \centering
    \caption{Complexity analysis of baselines and our method.}
    \scalebox{0.85}{
    \begin{tabular}{c|c|c}
    \toprule
              &   GCC                &   GCC+GraphControl \\ \midrule
\makecell{Forward\\ Time}  &  $LED+(L-1)ND^2+NKD$ & \makecell{$2(LED+(L-1)ND^2+NKD)$\\$+NK+NK^2+ND+ND^2$}  \\ \midrule
\makecell{Forward\\ Space} &  $E+(L-1)D^2+KD+LND$ & \makecell{$E+2((L-1)D^2+KD+LND)$\\$+NK+ND$} \\ \midrule
\makecell{Backward\\ Time} &  $LED+(L-1)ND^2+NKD$ & \makecell{$LED+(L-1)ND^2+NKD$\\$+NK+NK^2+ND+ND^2$} \\ \midrule
\makecell{Backward\\ Space} &  $E+(L-1)D^2+KD+LND$ & \makecell{$E+(L-1)D^2+KD+LND$\\$+NK+ND$} \\ \bottomrule
    \end{tabular}}
    \label{tab:complexity}
\end{table}

For the sake of clarity, we omit the smaller variables. Thus, the time complexity of GCC and our method can be simplified to:
\begin{table}[htp]
    \centering
    \caption{Complexity analysis of baselines and our method.}
    \scalebox{0.85}{
    \begin{tabular}{c|c|c}
    \toprule
              &   GCC                &   GCC+GraphControl \\ \midrule
\makecell{Time\\ Complexity}  &  $LED+(L-1)ND^2+NKD$  & \makecell{$LED+(L-1)ND^2+NKD$\\$+NK+NK^2+ND+ND^2$}  \\ \midrule
\makecell{Space\\ Complexity} &  $E+(L-1)D^2+KD+LND$ & \makecell{$E+(L-1)D^2+KD+LND$\\ $+NK+ND$}\\ \bottomrule
    \end{tabular}}
    \label{tab:total_complexity}
\end{table}

Compared to GCC, the additional term in our method's time and space complexity arises from two small zero MLPs. Consequently, our overall time and space complexity remains similar to GCC.

Furthermore, we list the empirical results of running time and memory cost during training:

\begin{table}[htp]
    \centering
    \caption{Running time of baselines and our method.}
    \begin{tabular}{c|c|c|c|c}
    \toprule
epoch/s&  Cora\_ML   & Photo  & DBLP  & Physics \\ \midrule
GCC    &   0.13      &  0.25   & 0.52   & 1.57     \\
GCC+GraphControl&0.15&  0.29   & 0.56   & 1.70     \\ \bottomrule
    \end{tabular}
    
    \label{tab:time}
\end{table}

For memory consumption, GCC and our method are also similar across different datasets:

It has to note that our method are conducted on subgraphs, so on different datasets, the memory consumption is similar, and it is memory efficient compared to transductive methods.

\begin{table*}[htp]
    \centering
    \caption{Statistics of pre-training datasets.}
    \begin{tabular}{l|cccccc}
    \toprule
     Dataset & Academia & DBLP (SNAP) & DBLP (NetRep) & IMDB & Facebook & LiveJournal \\ \midrule
    \#Nodes  & 137,969 & 317,080 & 540,486 & 896,305 & $3,097,165$ & $4,843,953$ \\
    \#Edges & 739,384 & $2,099,732$ & $30,491,458$ & $7,564,894$ & $47,334,788$ & $85,691,368$ \\
    \bottomrule
    \end{tabular}
    
    \label{tab:pretraining_data}
\end{table*}

\begin{table}[htp]
    \centering
    \caption{Memory consumption of baselines and our method.}
    \begin{tabular}{c|c|c|c|c}
    \toprule
memory/MB &  Cora\_ML    & Photo   & DBLP   & Physics \\ \midrule
GCC    &   426      &  478    & 420   & 458   \\
GCC+GraphControl& 428&  480   & 424   & 458   \\
\bottomrule
    \end{tabular}
    \label{tab:memory}
\end{table}

\section{Experiment}\label{app:experiment}
In this section, we will provide detailed information about experiments. Firstly, we introduce the datasets used in the main content in detail. And then we introduce the baselines used in the main content. Lastly, we provide the hyper-parameters of experiments.
\subsection{Datasets}\label{app:datasets}
\subsubsection{Pretraining datasets} 
The pre-training datasets utilized by GCC are outlined in Table~\ref{tab:pretraining_data}. These datasets fall into two main categories: academic graphs, including Academia, and two DBLP datasets, and social graphs, including IMDB, Facebook, and LiveJournal datasets. The Academia dataset is sourced from NetRep~\cite{ritchie2016scalable}, and the two DBLP datasets are obtained from SNAP~\cite{backstrom2006group} and NetRep~\cite{ritchie2016scalable} respectively. Additionally, the IMDB and Facebook datasets are gathered from NetRep~\cite{ritchie2016scalable}, and the LiveJournal dataset is collected from SNAP~\cite{backstrom2006group}.

\subsubsection{Downstream datasets}
The datasets can be categorized into two groups: attributed datasets (Cora\_ML, Amazon Photo, DBLP, and Coauthor Physics) and non-attributed datasets (USA Airport, Europe Airport, Brazil Airport, and H-index). Below are detailed descriptions of these datasets:
\begin{itemize}
\item Amazon Photo~\cite{coauthor} consists of segments from the Amazon co-purchase graph~\cite{mcauley2015image}. In this dataset, nodes represent goods, edges signify frequent co-purchases between goods, node features are bag-of-words encoded product reviews, and class labels are assigned based on product categories.
\item Cora\_ML and DBLP datasets~\cite{cora_ml} are citation networks used for predicting article subject categories. In these datasets, graphs are created from computer science article citation links. Nodes represent articles, and undirected edges signify citation links between articles. Class labels are assigned based on paper topics.
\item The Coauthor Physics dataset~\cite{coauthor} consists of co-authorship networks derived from the Microsoft Academic Graph. Nodes in this dataset represent authors, connected by edges if they co-authored a paper. Node features denote paper keywords from each author's publications, while class labels indicate the authors' most active fields of study.
\item The USA Airport dataset~\cite{struc2vec} consists of data collected from the Bureau of Transportation Statistics\footnote{https://transtats.bts.gov/} between January and October 2016. The network comprises 1,190 nodes and 13,599 edges, with a diameter of 8. Airport activity is quantified by the total number of people who passed through the airport (both arrivals and departures) during the corresponding period.
\item The Europe Airport dataset~\cite{struc2vec} comprises data gathered from the Statistical Office of the European Union (Eurostat)\footnote{http://ec.europa.eu/} between January and November 2016. The network consists of 399 nodes and 5,995 edges, with a diameter of 5. Airport activity is evaluated based on the total number of landings and takeoffs during the corresponding period.
\item The Brazil Airport dataset~\cite{struc2vec} is from the National Civil Aviation Agency (ANAC)\footnote{http://www.anac.gov.br/} and covers the period from January to December 2016. The network comprises 131 nodes and 1,038 edges, with a diameter of 5. Airport activity is quantified based on the total number of landings and takeoffs during the corresponding year.
\item The H-index dataset \cite{gcc} is derived from a co-authorship graph extracted from OAG\cite{oag}. Smaller subgraphs are extracted from the original graph due to its vast scale. This resulting network comprises 5,000 nodes and 44,020 edges, with a diameter of 7. Labels in H-index dataset indicate whether the author's h-index is above or below the median.
\end{itemize}

\subsection{Hyper-parameters}\label{app:hyper}
In this section, we will provide the hyper-parameters used in our experiments. Table~\ref{tab:baseline} lists the parameters of baselines. 
And Table~\ref{tab:transfer_app} provides the details of transfer learning.
\begin{table*}[htp]
\caption{Hyper-parameters for GIN(A,X) baseline.}
    \centering
    \begin{tabular}{c|cccccccc}
    \toprule
                  & Cora\_ML & Amazon-Photo & DBLP & Coauthor-Physics & USA & Europe & Brazil & H-index \\ \midrule
    Model         & GIN      & GIN          & GIN  & GIN              & GIN & GIN    & GIN    & GIN     \\
    \# Hidden size& 64       & 64           & 64   & 64               &     64&        64&        64&          64\\
    \# Layers     & 4        & 4            & 4    & 4                &     4&        4&        4&          4\\
    \# Epochs     & 1000     & 800          & 100  & 100              &     100&        100&        200&          200\\
    Learning rate & 1e-3     & 1e-2         & 1e-3 & 1e-3             &     1e-3&        1e-2&        1e-2&          1e-3\\
    Optimizer     & Adam     & Adam         & Adam & Adam             &     Adam&        Adam&        Adam&          Adam\\
    Weight decay  & 5e-4     & 5e-4         & 5e-4 & 5e-4             &     5e-4&        5e-4&        5e-4&          5e-4\\ \bottomrule
    \end{tabular}
    \label{tab:baseline}
\end{table*}

\begin{table*}[htp]
\caption{Hyper-parameters for Transfer Learning (\ours with GCC pre-trained model).}
    \centering
    \begin{tabular}{c|cccccccc}
    \toprule
                  & Cora\_ML & Amazon-Photo & DBLP & Coauthor-Physics & USA & Europe & Brazil & H-index \\ \midrule
    Model         & GCC      & GIN          & GCC  & GCC              & GCC & GCC    & GCC& GCC\\
    \# Hidden size& 64       & 64           & 64   & 64               & 64  & 64     &        64&          64\\
    \# Layers     & 4        & 4            & 4    & 4                & 4   & 4      &        4&          4\\
    \# Epochs     & 100      & 100          & 100  & 100              & 100 & 100    &        400&          100\\
    Learning rate & 0.5      & 0.5            & 0.1  & 0.01             & 0.3 & 0.2    &        0.1&          0.1\\
    Optimizer     & AdamW    & AdamW        & Adam & Adam             & SGD & SGD    &        SGD&          SGD\\
    Weight decay  & 5e-4     & 5e-4         & 5e-4 & 1e-2             & 1e-3& 5e-4   &        1e-3&          5e-4\\ 
    Walk steps    & 256      & 256          & 256  & 256              & 256 & 256    &        256&          256\\ 
    Restart rate  & 0.8      & 0.8          & 0.8  & 0.8              & 0.5 & 0.5    &        0.3&          0.5\\ 
    Threshold     & 0.17     & 0.2          & 0.3  & 0.15             & 0.15& 0.15   &        0.3&          0.17\\ 
    \bottomrule
    \end{tabular}
    \label{tab:transfer_app}
\end{table*}

\begin{table*}[htp]
\caption{Hyper-parameters for Domain Transfer (\ours with other pre-trained models).}
    \centering
    \begin{tabular}{c|cccccccc}
    \toprule
                  & Cora\_ML & Amazon-Photo & DBLP & Coauthor-Physics & USA & Europe & Brazil & H-index \\ \midrule
    Model         & GIN      & GIN          & GIN  & GIN              & GIN & GIN    & GIN    & GIN     \\
    \# Hidden size& 64       & 64           & 64   & 64               & 64  & 64     &        64&          64\\
    \# Layers     & 4        & 4            & 4    & 4                & 4   & 4      &        4&          4\\
    \# Epochs     & 100      & 100          & 100  & 100              & 100 & 100    &        200&          200\\
    Learning rate & 1e-1     & 1e-3         & 1e-3 & 1e-3             & 1e-3& 1e-3   &        1e-3&          5e-4\\
    Optimizer     & Adam     & Adam         & Adam & Adam             & Adam& Adam   &        Adam&          SGD\\
    Weight decay  & 1e-3     & 5e-4         & 5e-4 & 5e-4             & 5e-4& 5e-4   &        5e-4&          5e-4\\ 
    Walk steps    & 256      & 256          & 256  & 256              & 256 & 256    &        256&          256\\ 
    Restart rate  & 0.3      & 0.5          & 0.3  & 0.3              & 0.3 & 0.3    &        0.3&          0.3\\ 
    \bottomrule
    \end{tabular}
    \label{tab:my_label}
\end{table*}

\subsection{Baselines}\label{app:baseline}
In Section~\ref{sec:transfer_exp}, four pre-training methods are incorporated: GCC, GRACE, simGRACE, RoSA, and COSTA. In this section, we will elucidate these methods:
\begin{itemize}
\item GCC~\cite{gcc} is a structural pre-training method based on local structural information. It utilizes position embeddings as model input to learn transferable structural patterns through subgraph discrimination.
\item GRACE~\cite{grace} is a node-node graph contrastive learning method that employs augmentation functions (\ie, removing edges and masking node features) to create two augmented views. Then, InfoNCE~\cite{cpc} loss is applied to these views to maximize the lower bound of mutual information between positive views.
\item simGRACE~\cite{xia2022simgrace} eliminates data augmentation while introducing encoder perturbations to generate distinct views for graph contrastive learning.
\item RoSA~\cite{rosa} is a robust self-aligned graph contrastive framework that doesn't necessitate explicit node alignment in positive pairs, allowing for more flexible graph augmentation. It introduces the graph earth move distance (g-EMD) to compute the distance between unaligned views, achieving self-alignment. Additionally, it employs adversarial training for robust alignment.
\item COSTA~\cite{costa} is another graph contrastive learning method which proposes feature augmentation to decrease the bias introduced by graph augmentation.
\end{itemize}

\end{document}